\documentclass[conference]{IEEEtran}
\IEEEoverridecommandlockouts
\usepackage{cite}
\usepackage{amsmath,amssymb,amsfonts}
\usepackage{textcomp}

\usepackage{graphicx}
\usepackage{booktabs}
\usepackage{amsmath}
\usepackage{amsfonts}
\usepackage{algorithm}
\usepackage{algpseudocode}
\usepackage{array}

\usepackage[utf8]{inputenc} 
\usepackage[T1]{fontenc}    
\usepackage{hyperref}       
\usepackage{url}            
\usepackage{booktabs}       
\usepackage{amsfonts}       
\usepackage{nicefrac}       
\usepackage{microtype}      
\usepackage{xcolor}         

\def\BibTeX{{\rm B\kern-.05em{\sc i\kern-.025em b}\kern-.08em
    T\kern-.1667em\lower.7ex\hbox{E}\kern-.125emX}}
\begin{document}

\title{DMPCN: Dynamic Modulated Predictive Coding Network with Hybrid Feedback Representations}

\author{\IEEEauthorblockN{A S M Sharifuzzaman Sagar}
\IEEEauthorblockA{\textit{Department of Artificial Intelligence and Robotics} \\
\textit{Sejong University}\\
Seoul, South Korea \\
sharifsagar80@sejong.ac.kr}
\and
\IEEEauthorblockN{Yu Chen}
\IEEEauthorblockA{\textit{AI Lab} \\
\textit{MetaSyntec}\\
yuchen@metasyntec.com}
\and
\IEEEauthorblockN{Jun Hoong Chan}
\IEEEauthorblockA{\textit{School of Computer Science} \\
\textit{Peking University}\\
 Beijing, China \\
 junhoong95@stu.pku.edu.cn}
}

\maketitle

\begin{abstract}
Traditional predictive coding networks, inspired by theories of brain function, consistently achieve promising results across various domains, extending their influence into the field of computer vision. However, the performance of the predictive coding networks is limited by their error feedback mechanism, which traditionally employs either local or global recurrent updates, leading to suboptimal performance in processing both local and broader details simultaneously. In addition, traditional predictive coding networks face difficulties in dynamically adjusting to the complexity and context of varying input data, which is crucial for achieving high levels of performance in diverse scenarios. Furthermore, there is a gap in the development and application of specific loss functions that could more effectively guide the model towards optimal performance. To deal with these issues, this paper introduces a hybrid prediction error feedback mechanism with dynamic modulation for deep predictive coding networks by effectively combining global contexts and local details while adjusting feedback based on input complexity. Additionally, we present a loss function tailored to this framework to improve accuracy by focusing on precise prediction error minimization. Experimental results demonstrate the superiority of our model over other approaches, showcasing faster convergence and higher predictive accuracy in CIFAR-10, CIFAR-100, MNIST,
and FashionMNIST datasets.
\end{abstract}


\section{Introduction}
\label{sec:intro}
Predictive coding network \cite{PCN} is a computational model utilizing feedback mechanisms to generate predictions about sensory input, facilitating efficient processing of information in the brain. It is inspired by the predictive coding theory \cite{PCN}, which suggests that the brain predicts sensory inputs through hierarchical generative models and emerges as an alternative approach that aligns closely with biological mechanisms of human learning and perception. Unlike traditional convolutional networks \cite{deep} that rely on backpropagation, predictive coding networks employ a biologically plausible mechanism that iteratively adjusts based on prediction errors at each layer, potentially offering more efficient computation and a closer alignment with natural neural processes. This achieves a very useful potential computational efficiency, reducing reliance on extensive data and computations as compared to typical convolutional networks.

The prediction error feedback mechanism in conventional predictive coding networks either focus on local feature \cite{local} or global feature information processing \cite{global}, which limits the performance of the predictive coding networks. Moreover, they often exhibit limited contextual adaptability, failing to dynamically tailor processing strategies to the variations of inputs, which can hinder performance in diverse and dynamic visual environments. These limitations highlight the need for variants of predictive coding network models that can effectively navigate the complexities of image recognition by enhancing detail discernment and contextual responsiveness.

Recent advancements in PCNs for image recognition have predominantly focused on leveraging global recurrent update mechanisms to improve model performance, even with a reduced layer count \cite{global}. Concurrently, alternative strategies emphasize the utilization of local recurrent updates as a means to refine model efficacy. There is a notable gap in the existing literature in introducing a unified framework that integrates both global and local recurrent mechanisms for image recognition. Moreover, there is a noticeable oversight in the adaptability of these models, with a lack of context-aware dynamic modulation to tailor the network processing in response to specific input characteristics. Additionally, conventional approaches by applying standard loss functions neglect the potential benefits of employing loss functions specifically designed for predictive coding networks, which could further optimize image recognition performance. This highlights the need for a more integrated and contextually adaptive model guided by specifically tailored loss functions.

In this paper, we propose DMPCN, a \textbf{D}ynamic \textbf{M}odulated \textbf{P}redictive \textbf{C}oding \textbf{N}etwork with hybrid feedback representation by combining both global and local recurrent update to capture broad contextual information with fine-grained pixel details of the input images. DMPCN enables the network to adapt its processing based on the contextual information present within the input images, thereby enhancing the flexibility of the model and responsiveness to varied input images. Furthermore, we propose a loss function explicitly for predictive coding networks to optimize the performance of the network by more effectively leveraging the unique characteristics of predictive coding, focusing on minimizing prediction errors in a manner that directly correlates with improved image recognition accuracy. In summary, DMPCN not only addresses the limitations of traditional predictive coding network architectures but also shows the capability of the model in image recognition with smaller layers via the following contribution:
\begin{itemize}
    \item we propose a hybrid prediction error mechanism that combines local and global feedback within the network architecture;
    \item we introduce a dynamic modulation mechanism with easy adaptation of the input data; 
    \item we design a predictive consistency loss function tailored for the hybrid predictive coding network.
\end{itemize}
The experimental results indicate that our proposed DMPCN achieves superior performance compared to other object recognition approaches.







\section{Related Work}\label{sec:related}
This section mainly discusses predictive coding and recurrent processing approaches in predictive coding networks, such as local and global updates.
\subsection{Predictive Coding} 
Predictive Coding is an important concept in understanding the information-processing mechanisms of the human brain~\cite{RW-1}. The fundamental of predictive coding lies in the ability of the brain to learn how to reduce the prediction errors across neurons by utilizing an internal generative model to anticipate incoming stimuli~\cite{RW-2}. This framework not only approximates but, under certain conditions, can also mirror the weight update dynamics of backpropagation in layered networks~\cite{RW-3}, and is further generalized across different models~\cite{RW-4}. Notably, the weight update of backpropagation can be replicated by predictive coding precisely with the integration of external controls~\cite{RW-5}, encompassing the flexibility of predictive coding in modalities during training and testing phases, its intricate mathematical underpinnings, and its formulation as an energy-based model \cite{RW-6,RW-7,RW-8}. The research scope of predictive coding extends to potentially unifying brain function theory, providing a comprehensive framework for understanding perception, learning, and cognitive neural substrates.
\subsection{Local Recurrent Processing} 
Local Recurrent Processing is pivotal in the object recognition capabilities of the brain, with studies emphasizing both feedforward mechanisms for rapid recognition~\cite{RW-9} and feedback connections for top-down predictions~\cite{RW-10}. The feedback signals operate between hierarchically adjacent areas and are essential for disambiguating visual inputs, especially under ambiguous conditions, enhancing object recognition through local recurrent processing~\cite{RW-11}. Furthermore, the integration of top-down attention and bottom-up sensory information through global and local processing is crucial for efficient recognition, facilitated by feedback processes that coordinate information across the brain \cite{RW-13, Rw-14}. Advanced neuroimaging techniques have shed light on these dynamic interactions and the critical role of feedback in modulating neural activity for recognition \cite{RW-15}. These findings highlight the need for ongoing research to understand the intricate neural mechanisms behind object recognition with local recurrent processing.

\subsection{Global Recurrent Processing.} 

The introduction of a predictive coding network incorporating global recurrent processing, as outlined in \cite{global}, signifies a significant advancement. This model seamlessly integrates feedforward and feedback computations throughout its architecture, facilitating comprehensive information integration via recurrent dynamics. Nonetheless, the global recurrent processing approach, may not fully optimally address the demands of image recognition tasks that necessitate extensive localized information processing.

The current research within predictive coding networks often restricts its scope to particular feedback mechanisms, which may not fully exploit the complexities of image recognition tasks. To address this issue, we propose a hybrid approach that combines local and global feedback within the architecture to enhance the ability of network to process information at both macro and micro levels. The hybrid model is further improved with a dynamic modulation mechanism, enabling the network to adapt its processing in response to the context and specificities of the input data.



\section{Problem Formulation}
\label{sec:problem}
This section begins with the introduction of a predictive coding network, focusing on the error prediction mechanism, including the learning dynamics of the feedforward and feedback processes. Subsequently, we describe the problem of predictive coding networks and how they can be formulated using local and global recurrent processing.

\subsection{Introduction to Predictive Coding Network}
The main component of a predictive coding network is the error signal generated from the difference between the predicted sensory input and the actual sensory data received. The goal is to minimize these error signals and continuously update the prediction based on the dynamic learning mechanism and it is crucial for the refinement of future prediction.


The mechanism for generating predictions is captured by the interaction between the current layer $l$ and the previous layer $l-1$ and is computed as
\begin{equation} \label{Eq:1_prediction}
\mathbf{p}_{l-1}(t)=\mathbf{W}_{l, l-1}^T  \mathbf{r}_l(t),
\end{equation}
where $\mathbf{p}_{l-1}(t)$ is the prediction for layer $l-1$ at time $t$, $\mathbf{W}_{l, l-1}^T$ is the transpose of the weight matrix from layer $l$ to $l-1$, and $\mathbf{r}_l (t)$ is the representation of the actual sensory input in layer $l$ at time $t$.

The prediction error in the predictive coding network serves as a feedback signal to refine the model weight to ensure a dynamic adaptation to the sensory environment. Given the representation $\mathbf{r}_{l-1} (t)$ from the previous layer and the prediction $\mathbf{p}_{l-1}(t)$ for the current layer, the prediction error $\mathbf{e}_{l-1}(t)$ can be calculated as
\begin{equation}\label{Eq:2_error}
\mathbf{e}_{l-1}(t)=\mathbf{r}_{l-1}(t)-\mathbf{p}_{l-1}(t).
\end{equation}

The dynamic learning mechanism of the predictive coding network consists of a feedforward and feedback process. The feedforward process utilizes the prediction error in Eq.~\ref{Eq:2_error} to update the representation in the next layer to minimize the future prediction difference. The loss function of the feedforward mechanism is the squared norm of the prediction error, normalized by the variance of the errors, and it can be defined as follows:
\begin{equation} \label{Eq:forward_loss}
L_{l-1}(t)=\frac{1}{2 \sigma_{l-1}^2}\left\|\mathbf{e}_{l-1}(t)\right\|^2 ,
\end{equation}
the calculation of the gradient of the loss function in Eq. \ref{Eq:forward_loss} to the layer representation $r_l (t)$ helps us understand what changes need to be made to decrease the prediction error. This can be expressed as 
\begin{equation} \label{Eq:forward_gradient}
\frac{\partial L_{l-1}(t)}{\partial \mathbf{r}_l(t)}=-\frac{1}{\sigma_{l-1}^2} \mathbf{W}_{l, l-1} \mathbf{e}_{l-1}(t).
\end{equation}
Thus, the representation $r_l (t)$ is updated by a gradient descent step, factoring in the learning rate $\alpha$, as follows:
\begin{equation} \label{Eq:forward_representation}
\mathbf{r}_l(t+1)=\mathbf{r}_l(t)-\alpha \frac{\partial L_{l-1}(t)}{\partial \mathbf{r}_l(t)}.
\end{equation}



In the feedback process, the differences between the predicted and the actual sensory inputs at each layer are utilized, to iteratively update the representations. For any given layer $l$, the prediction error, $\mathbf{e}_l (t)$, is calculated as the difference between the actual representation at layer $l$, $\mathbf{r}_l (t)$, and the predicted representation for the same layer, $\mathbf{p}_l (t)$. The correction of the representation $\mathbf{r}_l (t)$ in response to this prediction error is directed by the gradient of the prediction error for $\mathbf{r}_l (t)$. This can be formulated as 
\begin{equation}\label{Eq:backward_gradient}
\frac{\partial L_l(t)}{\partial \mathbf{r}_l(t)}=\frac{2}{\sigma_l^2}\left(\mathbf{r}_l(t)-\mathbf{p}_l(t)\right),
\end{equation}
where $\sigma_l^2$ represents the variance of the prediction error. A gradient descent step is then utilized to update the representations using the following equation,
\begin{equation}\label{Eq:backward_representation}
\mathbf{r}_l(t+1)=\mathbf{r}_l(t)-\beta_l \frac{\partial L_l(t)}{\partial \mathbf{r}_l(t)}=\left(1-\frac{2 \beta_l}{\sigma_l^2}\right) \mathbf{r}_l(t)+\frac{2 \beta_l}{\sigma_l^2} \mathbf{p}_l(t).
\end{equation}



\subsection{The Recurrent Processing in Predictive Coding Network}
The common recurrent processing in predictive coding networks either focuses on contextual information through global recurrent cycles or enhances detail processing via local recurrent cycles for object recognition tasks. These recurrent cycles can lead to suboptimal performance in complex visual scenes. Moreover, it lacks the mechanism to dynamically adjust its processing based on the complexity, scale, and context of the input image, which is crucial when dealing with high variability environments. The process of global and local recurrent cycles is defined as follows:
\begin{equation} \label{Eq:recurrent_processing}
\begin{aligned}
    E_{global} &=\sum_{l=1}^L\left(\left\|R_l(X_{global})-P_{l \rightarrow l+1}\left(R_{l+1}(X_{global})\right)\right\|_2^2\right),\\
    E_{local } &=\sum_{l=1}^L\left(\left\|R_l\left(X_{local}\right)-P_l\left(R_l\left(X_{local }\right)\right)\right\|_2^2\right),
\end{aligned}
\end{equation}
where $R_l(X_{global})$ and $R_l (X_{local})$ denote as the actual representations at layer $l$ for global and local contexts, respectively. $P_{l \rightarrow l+1}$ is the prediction functions from layer $l$ to $l+1$ and vice versa for global interactions, $P_l$ signifies the prediction function within the same layer $l$ for local adjustments, and $X_{local}$  represents localized features within the input $X$. 

Eq.~\ref{Eq:recurrent_processing} highlights the paradox between processing global contextual information and focusing on local details, each with its shortcomings. The global approach may overlook crucial local features, while the local method may miss broader contextual cues necessary for accurate object recognition. Moreover, traditional PCN cannot adapt its processing to the variability of image context, and the function can be expressed as
\begin{equation}
E_{\text {static }}=\sum_{l=1}^L\left[(]\phi\left(R_l(X), C\right)\right],
\end{equation}
where $\phi$ represents a function measuring the inability of the network to adapt to the context $C$ inherent in the input $X$.


\section{Proposed Method}
\label{sec:method}
\begin{figure}[t]
    \includegraphics[width=1.0\linewidth]{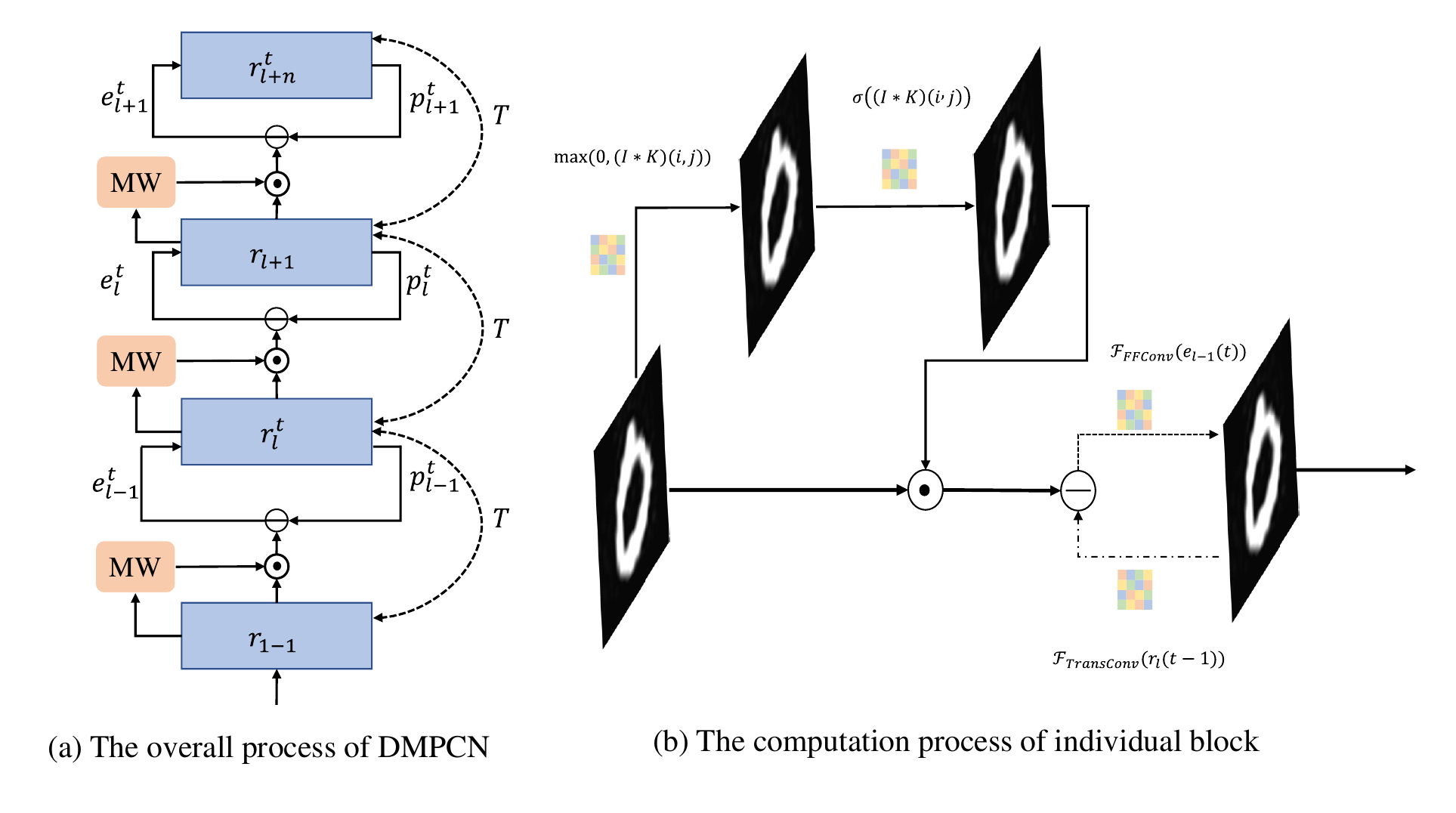}
    \caption{Dynamic Modulated Predictive Coding Network~(DMPCN). (a) the overall process of our proposed DMPCN, and (b) the computation process of individual blocks.}
    \label{fig:1}
\end{figure}

In this section, we introduce the DMPCN as shown in Fig.~\ref{fig:1}, where the local recurrent update, global recurrent update, dynamic modulation mechanism, and predictive consistency loss function are specifically designed for complex object detection scenes.

\subsection{Local Recurrent Update}
Motivated by \cite{local}, we introduce a local recurrent update method to enhance the capability of PCN for detailed feature extraction and error correction at a granular level by combining standard convolutional and transpose convolutional operations. Given an input feature map at layer $l$, denoted as $x_l \in \mathbb{R}^{H \times W \times C_{i n}}$, where $H, W$, and $C_{i n}$ represent the height, width, and number of input channels, respectively. The input feature map is then passed through the convolutional layer with kernel size $k$, stride $s$, and padding $p$, which can be represented as:
\begin{equation} \label{Eq:method_local_1}
y_l=\operatorname{ReLU}\left(W_{f f} * x_l\right),
\end{equation}
where $W_{f f} \in \mathbb{R}^{k \times k \times C_{i n} \times C_{o u t}}$ denotes the weights of the feedforward convolution, $y_l$ is the output feature map, and $\ast$  denotes the convolution operation.
The feedback pass involves a convolutional transpose operation intended to project the feature map back to the input space of layer $l$ represented as follows:
\begin{equation} \label{Eq:method_local_2}
y_l^{\prime}=W_{f b} \star y_l,
\end{equation}
where $W_{f b}=W_{f f}^T$, $\star$ denotes the convolutional transpose operation, and $y_l^{\prime}$ is the feedback projection of $y_l$. Then the feedback error correction is done using the following function:
\begin{equation} \label{Eq:method_local_3}
y_{adj}=\operatorname{ReLU}\left(\beta_0 \odot\left(x_l-y_l^{\prime}\right)+y_l\right),
\end{equation}
where $\beta_0$ is the learnable parameter that is modulated through a non-linear activation function, $\odot$ denotes element-wise multiplication.

\subsection{Global Recurrent Update}
The global recurrent update is primarily responsible for employing both feedforward and feedback paths across the network hierarchy to minimize prediction error as shown in Eq.~\ref{Eq:2_error}. The feedback signal of layer $l$ derived from the prediction error at the previous layer $l+1$ is propagated back as follows:
\begin{equation}\label{Eq:global_e}
\textbf{e}_l^{\text {feedback }}=\operatorname{ConvTranspose 2 \mathrm{D}}\left(\textbf{e}_{l+1} ; W_{f b}^{(l)}\right),
\end{equation}
where $W_{f b}^{(l)}$ represents the weights used in the feedback path from layer $l+1$ to $l$.
Subsequently, the global recurrent update for the representation at layer $l$, $x_l^{\prime}$ incorporates the feedback error $e_l^{\text {feedback }}$ to adjust $x_l$ under the update rate $\alpha$, which can be represented as
\begin{equation}\label{Eq:global_xprime}
x_l^{\prime}=\operatorname{ReLU}\left(x_l+\alpha_l \cdot \textbf{e}_l^{\text {feedback }}\right).
\end{equation}

\subsection{Dynamic Modulated Predictive Coding Network (DMPCN)} 
We propose a DMPCN to compute the modulation factors based on the error signals (as shown in Eq.~\ref{Eq:2_error}). This allows the feedback signals on the network representation to be adjusted dynamically. For layer $l$, the modulation factor $m_l$ can be defined as follows:
\begin{equation} \label{Eq:dynamic_factor}
m_l=\sigma\left(\operatorname{Conv2\mathrm{D}} \left(\textbf{e}_l ; W_{m o d}^{(l)}\right)\right),
\end{equation}
where $W_{\bmod }^{(l)}$ are the weights of the modulation layer at layer $l$ and $\sigma$ is the sigmoid function. Next, the modulation factor in Eq.~\ref{Eq:dynamic_factor} is then used to dynamically update the representation $x_l^{\prime}$ during the recurrent update process as follows: 
\begin{equation}\label{Eq:dynamic_x_1}
x_l^{\prime}=x_l+\left(m_l \odot \textbf{e}_l^{\text {feedback }}\right),
\end{equation}
where $\odot$ represents the element-wise multiplication. This modulation factor $m_l$ scales the feedback error $e_l^{\text {feedback }}$ (as shown in Eq.~\ref{Eq:global_e}) before updating the representation $x_l$ to allow the network to focus on more relevant signals.

\subsection{Predictive Consistency Loss Function}
We specifically design a predictive consistency loss~(PCL) function to guide our proposed method in acquiring optimal performance in object recognition tasks. The loss function consists of four individual parts: hybrid loss, spatial consistency term, spatial loss, and reconstruction loss.

\subsubsection{The Hybrid Loss.} We introduce a hybrid loss $L_{\text {hybrid}}$ which integrates the cross-entropy loss using a dynamic modulation factor from the modulation weights of the network as follows:
\begin{equation} \label{Eq:loss_hybrid}
L_{\text {hybrid }}=\mathcal{L}_{C E}(O, Y) \times\left(1+\lambda \times \frac{1}{N} \sum_{i=1}^N M_i\right),
\end{equation}
where $\mathcal{L}_{C E}(O, Y)$ is the cross entropy loss computed between the output probabilities $O$ of the network and the ground truth label $Y$, $M_i$ represents the modulation weights and $\lambda$ is the scalar value to weights the dynamic modulation factor to integrate with the cross entropy loss.

\subsubsection{The Spatial Consistency Term} The spatial consistency term is introduced to evaluate the uniformity and stability across the spatially distributed features within the internal representations of the network. The spatial consistency term, denoted as $L_{S C T}$ is calculated across the total number of feature maps $N$ using the mean-variance of average activations across four quadrants which were derived from the normalized feature maps acquired from the network layers. 
It can be defined as follows:

\begin{equation}\label{Eq:loss_spatial_consistency}
\begin{aligned}
L_{\text{SCT}} 
&= \frac{1}{N} \sum_{i=1}^{N} 
   \operatorname{Var}\bigl(\operatorname{mean}(Q_{1i}), \operatorname{mean}(Q_{2i}),\\
&\qquad\qquad \operatorname{mean}(Q_{3i}), \operatorname{mean}(Q_{4i})\bigr).
\end{aligned}
\end{equation}


\subsubsection{The Spatial Loss.} We introduce spatial loss $L_{S P}$ to evaluate the capability of the network to maintain spatial coherence across hierarchical representations of our proposed model. The spatial consistency loss is defined as: 
 \begin{equation} \label{Eq:loss_spatial}
L_{S P}=\frac{1}{L} \sum_{l=1}^L \mathcal{L}_{M S E}\left(A_l, P_l\right),
\end{equation}
where $\mathcal{L}_{M S E}\left(A_l, P_l\right)$ is the mean squared error between the actual ($A_l$) and the predicted ($P_l$) representation at each layer $l$ across the of $L$ layers.

\subsubsection{The Reconstruction Loss.} We also introduce a reconstruction loss $L_{\text {recon }}$ to minimize the difference between the original input and its reconstructed counterpart to ensure the ability of the network to capture and regenerate the salient features of the input data. It can be defined as follows:
\begin{equation} \label{Eq:loss_recon}
L_{\text {recon }}=\mathcal{L}_{M S E}(X, \hat{X}),
\end{equation}
where $X$ is the original input and $\hat{X}$ is the reconstructed counterpart of the original input. 

By combining the loss function in Eq.~\ref{Eq:loss_hybrid}, Eq.~\ref{Eq:loss_spatial_consistency}, Eq.~\ref{Eq:loss_spatial}, and Eq.~\ref{Eq:loss_recon}. The overall loss function $L_{\text {total }}$for our proposed method is then defined as follows:
\begin{equation} \label{Eq:loss_total}
L_{\text {total }}= L_{\text {hybrid }} + \mu \cdot L_{S C T} + L_{S P} +\gamma \cdot L_{\text {recon }},
\end{equation}
where $\mu$ and $\gamma$ are the weighting factor for spatial and reconstructed loss.


\subsection{Datasets and Implementation}
\subsubsection{Datasets.} We evaluate the proposed model on three datasets, including CIFAR-10~\cite{krizhevsky2009learning}, CIFAR-100~\cite{krizhevsky2009learning}, and MNIST~\cite{lecun1998gradient}. The CIFAR-10~\cite{krizhevsky2009learning} and CIFAR-100~\cite{krizhevsky2009learning} datasets include $50,000$ training images and $10,000$ testing images in $10$ and $100$ categories, respectively. Each image is a $32\times32$ RGB image. Our proposed model is trained on the training set and evaluated on the test set.

\subsubsection{Implementation details.} We use the stochastic gradient descent~(SGD)~\cite{ruder2016overview} optimizer to train our proposed model with a momentum of $0.9$ and a weight decay of $5\times 10^{-4}$. We set the batch size to $64$ on all datasets for $250$ epochs. The initial learning rate is set to $1\times10^{-2}$ and is divided by $10$ at $80$, $140$, and $200$ epochs. All experiments are performed with PyTorch on NVIDIA RTX A$6000$ GPU with $48$~GB RAM.
\section{Experiments}
\label{sec:experiment}

Our experimental evaluation consists of three subsections: 1) recurrent cycle effect, 2) analysis of classification performance, and 3) comparison of different loss functions.


\subsection{Analysis of Classification Performance}
We demonstrate that DMPCN exhibits superior performance compared to the backpropagation~(BP) method and predictive coding network~(PCN) on multiple benchmarks. Our experiments evaluate the performance of DMPCN on various datasets, ensuring that all methods ran until convergence, with early stopping used to select the best-performing model.

In our experimental setup, we assess the image classification benchmarks using BP, PCN, and DMPCN. Initially, we train LeNet on the MNIST dataset, followed by the same network on FashionMNIST. For more complex tasks, we use AlexNet in CIFAR-10 and CIFAR-100. Finally, VGG9 was trained on both CIFAR-10 and CIFAR-100. Each model was rigorously tuned for hyperparameter optimization through grid search to ensure the highest test accuracy surpassing BP.
    
\begin{table}[ht]
\small
\setlength{\tabcolsep}{5pt}
    \footnotesize
    \centering
    \caption{Comparison of test accuracy on three different algorithms such as BP, PCN, and DMPCN on the CIFAR-10~\cite{krizhevsky2009learning}, CIFAR-100~\cite{krizhevsky2009learning}, MNIST~\cite{lecun1998gradient} and FashionMNIST~\cite{fashion} datasets. }\label{tab:2}
    \scalebox{0.9}{
    \begin{tabular}{lccc}
    
    \toprule
         & \textbf{BP} & \textbf{PCN} &\textbf{DMPCN} \\
         \hline
         LeNet on MNIST  &  \textbf{98.83\%}$\pm$0.42\%&  96.72\%$\pm$0.95\%& 98.78\%$\pm$0.87\% \\
         LeNet on FashionMNIST&  90.31\%$\pm$0.21\%&  84.53\%$\pm$0.37\%& \textbf{91.28\%}$\pm$0.64\% \\
         AlexNet on CIFAR10 &  83.29\%$\pm$0.88\% &  89.52\%$\pm$0.97\% &  \textbf{89.65\%}$\pm$0.53\% \\
         AlexNet on CIFAR100& 60.21\%$\pm$1.09\% & 58.25\%$\pm$0.43\% &\textbf{61.72\%}$\pm$0.75\% \\
         VGG9 on CIFAR10&  90.65\%$\pm$0.76\%& 93.83\%$\pm$0.51\% &\textbf{94.33\%}$\pm$0.38\% \\
         VGG9 on CIFAR100& 62.17\%$\pm$0.53\% &72.58\%$\pm$0.23\%  &  \textbf{74.84\%}$\pm$0.44\% \\
         \bottomrule
    \end{tabular} }
\end{table}

As shown in Table~\ref{tab:2}, DMPCN consistently outperforms PCN and BP, except for a marginal deficit in the simplest case (LeNet on MNIST), where BP leads by 0.05\%. PCN struggles to scale to more complex scenarios, being surpassed by all other methods. In contrast, DMPCN maintains stable performance across varying sizes, architectures, and datasets, often surpassing the BP method. 
\begin{figure}[ht]
    \centering
    \includegraphics[width=0.95\linewidth]{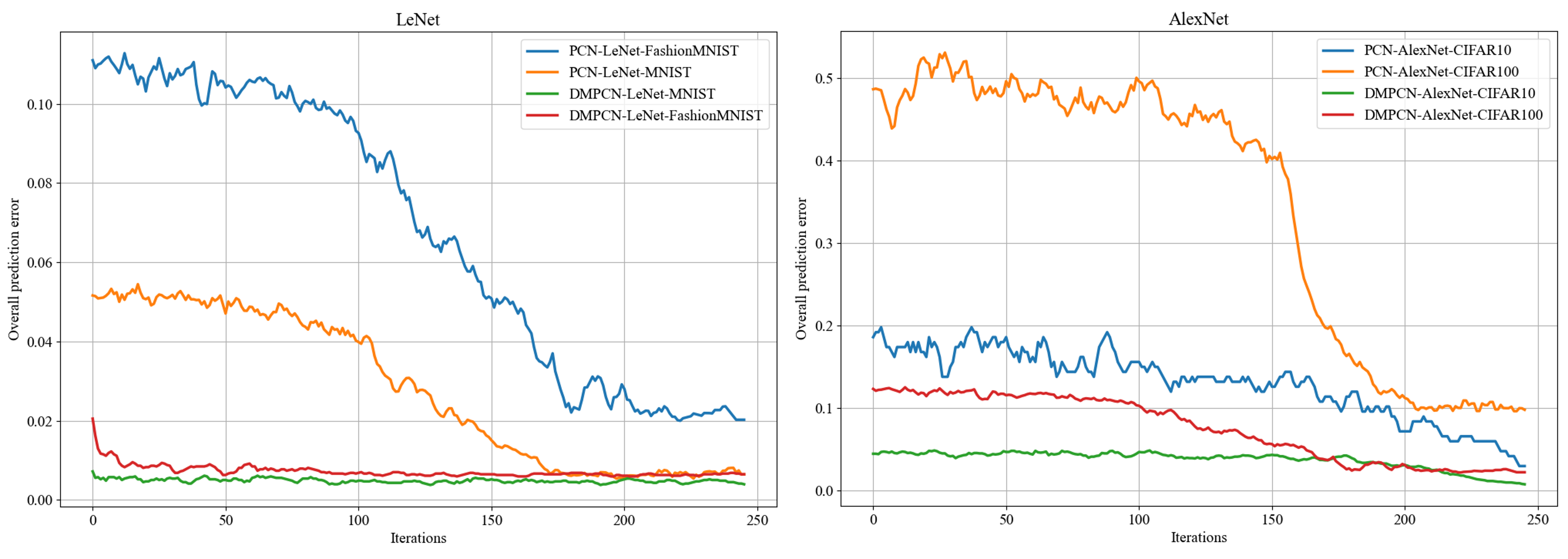}
    \caption{Comparison of PCN and DMPCN overall prediction error acquired by using Eq. \ref{Eq:loss_spatial}. (a) prediction error convergence comparison on MNIST and FashionMNIST datasets with PCN and DMPCN, respectively; (b)prediction error convergence comparison on CIFAR-10 and CIFAR-100 datasets with PCN and DMPCN, respectively.}
    \label{fig:prediction_error}
\end{figure}
We analyze the convergence of the overall prediction error for both the PCN and DMPCN methods. By using Eq.\ref{Eq:loss_spatial}, we calculate the prediction error for each batch and plot corresponding to the number of iterations. We employ the LeNet model with PCN and DMPCN methods on the MNIST and Fashion MNIST datasets, as shown in Fig.~\ref{fig:prediction_error}(a). DMPCN converges faster than the traditional PCN, attributed to the introduction of a global-local hybrid feedback and dynamic modulation mechanism. To further validate the results obtained with LeNet, we use the AlexNet model on the CIFAR-10 and CIFAR-100 datasets with both PCN and DMPCN methods, analyzing the convergence of the prediction error as shown in Fig.~\ref{fig:prediction_error}(b). These results similarly demonstrate that DMPCN converges faster compared to PCN. In general, these findings highlight the effectiveness of DMPCN in achieving faster convergence, whereas traditional PCN requires more iterations to converge.

\subsection{Calibration Analysis}



\begin{figure}[t]
    \centering
    \includegraphics[width=1.0\linewidth]{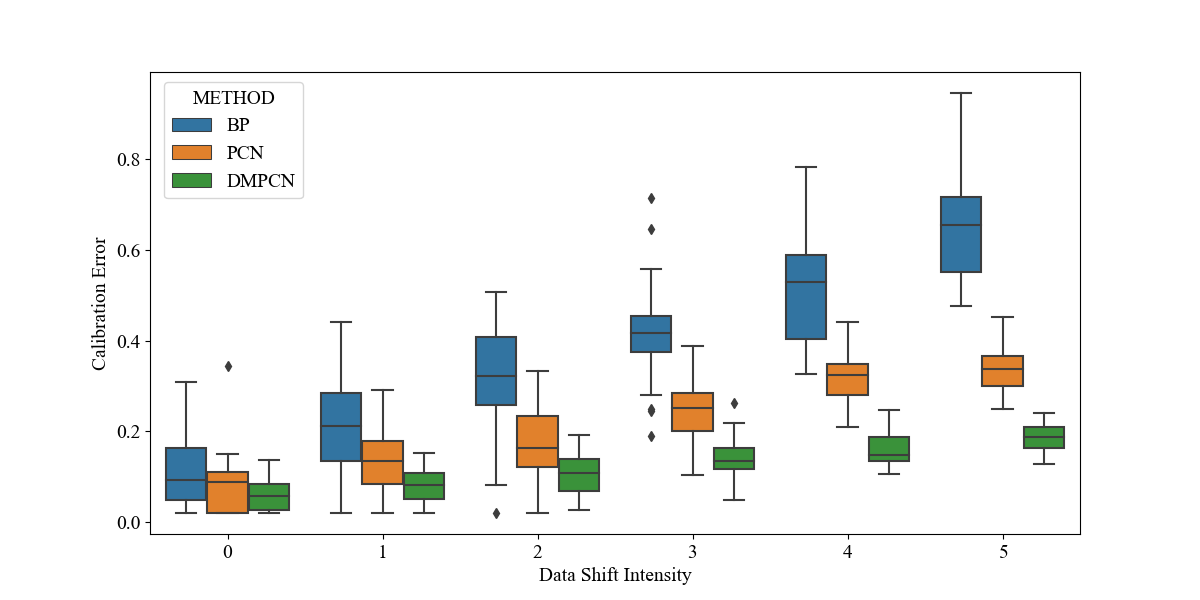}
    \caption{Calibration error analysis of BP, PCN and DMPCN on CIFAR10 dataset using VGG-9 across varying levels of data corruption, including rotation, Gaussian blur, Gaussian noise, hue, brightness, and contrast. The DMPCN model shows superior calibration compared to BP and PCN, particularly under different degrees of data shift, demonstrating its robustness in maintaining accurate model calibration.}
    \label{fig:3_calibration_err}
\end{figure}

Inspired by \cite{emde2024stable}, we analyze the calibration performance of BP, PCN, and DMPCN under various additive noises. Fig.~\ref{fig:3_calibration_err} illustrates the distributions of calibration errors for DMPCN, PCN, and BP at various levels of data corruption. The DMPCN model produces better-calibrated outputs, which is crucial for uncertainty quantification. In the in-distribution data, DMPCN shows a lower average calibration error than BP and PCN. Additionally, the increase in calibration error is significantly weaker for DMPCN as the intensity of the data shift increases. DMPCN consistently maintains lower median calibration errors across all levels of shift intensities, outperforming BP and PCN even up to level $5$ shifts. These results suggest that DMPCN is particularly well suited for applications where reliable uncertainty quantification is essential.

\subsection{Comparison of Loss Function}
Table.~\ref{tab:3} shows the error rate comparison on CIFAR-10~\cite{krizhevsky2009learning} and MNIST~\cite{lecun1998gradient} datasets using different loss functions. DMPCN combined with Kullback–Leibler divergence loss~\cite{kingma2013auto}, $L_{KL}$ achieved an error rate of $6.20$ on CIFAR-10~\cite{krizhevsky2009learning} and $0.45$ on the MNIST~\cite{lecun1998gradient} dataset. DMPCN with cross-entropy loss~\cite{lecun2002efficient}, $L_{CE}$ achieved an error rate of 6.35 on CIFAR-10~\cite{krizhevsky2009learning} and $0.36$ on the MNIST~\cite{lecun1998gradient} dataset. On the other hand, DMPCN with our proposed PCL loss achieved the lowest error rate of $5.67$ on CIFAR-10~\cite{krizhevsky2009learning} and $0.33$ on MNIST~\cite{lecun1998gradient} dataset.    

\begin{table}[th]
\setlength{\tabcolsep}{15pt}
    \centering
    \caption{Comparison of different loss functions in our proposed DMPCN on the CIFAR-10~\cite{krizhevsky2009learning} and MNIST~\cite{lecun1998gradient} datasets.}  \label{tab:3}
    \scalebox{0.9}{
    \begin{tabular}{ccc}
    \toprule
         \textbf{Method}& \textbf{CIFAR-10}~\cite{krizhevsky2009learning} & \textbf{MNIST}~\cite{lecun1998gradient} \\
         \hline
         DMPCN + $L_{KL}$ ~\cite{kingma2013auto} & 6.20& 0.45 \\
         DMPCN + $L_{CE}$ ~\cite{lecun2002efficient} & 6.35& 0.36\\
         DMPCN + $L_{PCL}$ (Ours)& \textbf{5.67} & \textbf{0.28}\\
         \bottomrule
    \end{tabular}}
\end{table}


\subsection{Ablation Study}
Table \ref{tab:my_label} presents the ablation study to evaluate the impact of various loss components, such as the Hybrid loss, Spatial Consistency Term (SCT) loss, Reconstruction~(Recons) loss, and Spatial loss, on the model's performance in the MNIST, CIFAR10, and CIFAR100 datasets. The baseline model with only the hybrid component achieved $99.56\%$, $93.35\%$, and $73.97\%$, respectively. The spatial consistency term loss component further improved the performance to $99.64\%$, $93.88\%$, and $74.54\%$. The reconstruction loss component achieved $99.66\%$, $93.85\%$, and $74.41\%$ accuracy. The spatial component alone results in 99.62\%, $94.04\%$, and $74.05\%$. Combining hybrid with spatial consistency term loss and reconstruction loss achieves $99.61\%$, $93.77\%$, and $73.98\%$. The hybrid with spatial consistency term loss and spatial loss combination acquired $99.63\%$, $93.65\%$, and $74.08\%$ accuracy. The integration of all components achieves the highest accuracy at $99.72\%$, $94.33\%$, and $74.84\%$.
      
     
  
       
       
        
     

\begin{table}[ht]
    \centering
    \caption{Ablation study of the performance of our loss components on MNIST, CIFAR10, and CIFAR100 datasets.}
    \renewcommand{\arraystretch}{1.2}
    \scalebox{0.8}{
    \begin{tabular}{c>{\centering\arraybackslash}p{1cm}cccccc}
     \toprule
        \textbf{Method} & \textbf{Hybrid} & \textbf{SCT} & \textbf{Recons} & \textbf{Spatial} & \textbf{MNIST} & \textbf{CIFAR10} & \textbf{CIFAR100} \\
        \hline
        &$\checkmark$ & - & - & - & 99.56 & 93.35 & 73.97 \\
      
        &$\checkmark$ & $\checkmark$ & - & - & 99.64 & 93.88 & 74.54 \\
     
        &$\checkmark$ & - & $\checkmark$ & - & 99.66 & 93.85 & 74.41 \\
  
        DMPCN&$\checkmark$ & - & - & $\checkmark$ & 99.62 & 94.04 & 74.05 \\
       
        &$\checkmark$ & $\checkmark$ & $\checkmark$ & - & 99.61 & 93.77 & 73.98 \\
       
        &$\checkmark$ & $\checkmark$ & - & $\checkmark$ & 99.63 & 93.65 & 74.08 \\
        
        &$\checkmark$ & - & $\checkmark$ & $\checkmark$ & 99.67 & 93.72 & 74.31 \\
     
        &$\checkmark$ & $\checkmark$ & $\checkmark$ & $\checkmark$ & 99.72 & 94.33 & 74.84 \\
        \bottomrule
    \end{tabular}   }
    \label{tab:my_label}
\end{table}

\section{Conclusion}
\label{sec:conclusion}
In this paper, we introduce a novel enhancement for traditional predictive coding networks through a hybrid prediction error feedback mechanism with dynamic modulation, effectively combining global and local contexts and adjusting feedback based on input complexity. Our approach, complemented by a specialized loss function designed for precise prediction error minimization, demonstrates superior predictive accuracy and faster convergence on benchmark datasets compared to existing methods. This work not only addresses key limitations in processing detailed and broader aspects of data and in adapting to varied input scenarios but also opens new avenues for future research in artificial intelligence and machine learning with a promising direction.
\bibliographystyle{abbrv}
\bibliography{main}
\end{document}